# A CONTEXT-SENSITIVE WORD EMBEDDING APPROACH FOR THE DETECTION OF TROLL TWEETS

Seyhmus YILMAZ[1.] And Sultan ZAVRAK[1*]

Seyhmus YILMAZ

[1]Department of Computer Engineering, Duzce University, Duzce/TURKEY

seyhmusyilmaz@duzce.edu.tr

***ORCID ID:*** 0000-0001-9987-2797

Sultan ZAVRAK

[1]Department of Computer Engineering, Duzce University, Duzce/TURKEY

sultanzavrak@duzce.edu.tr

***ORCID ID:*** 0000-0001-6950-8927

***Corresponding Author:*

Sultan ZAVRAK

Department of Computer Engineering,

Faculty of Engineering, Duzce University, 81620, Duzce / TURKEY

Phone: +90 (380) 542 1036

E-mail:  sultanzavrak@duzce.edu.tr

# A CONTEXT-SENSITIVE WORD EMBEDDING APPROACH FOR THE DETECTION OF TROLL TWEETS

**Abstract**

In this study, we aimed to address the growing concern of trolling behavior on social media by developing and evaluating a set of model architectures for the automatic detection of troll tweets. Utilizing deep learning techniques and pre-trained word embedding methods such as BERT, ELMo, and GloVe, we evaluated the performance of each architecture using various metrics. Our results indicate that BERT and ELMo embedding methods performed better than the GloVe method, likely due to their ability to provide contextualized word embeddings that better capture the nuances and subtleties of language use in online social media. Additionally, we found that CNN and GRU encoders performed similarly in terms of F1 score and AUC, suggesting their effectiveness in extracting relevant information from input text. The best-performing method was found to be an ELMo-based architecture that employed a GRU classifier, with an AUC score of 0.929. This research highlights the importance of utilizing contextualized word embeddings and appropriate encoder methods in the task of troll tweet detection, which can assist social-based systems in improving their performance in identifying and addressing trolling behavior on their platforms.

1. Introduction

The exponential growth and ubiquity of the Internet have resulted in the emergence of various social media environments that specialize in communication services (Tomaiuolo et al., 2020). Users in a diverse range of countries (Larosiliere et al., 2017) have adopted social media, in addition to firms (Meske, Junglas, & Stieglitz, 2019; Meske, Wilms, et al., 2019; Meske & Stieglitz, 2013), and the importance of "being social" has been acknowledged as crucial for the relationship quality between users in the online environment as opposed to exhibiting troll-like behavior (Chinnov et al., 2015; Meske, Junglas, Schneider, et al., 2019; Stieglitz et al., 2020). Trolls, who purport to desire to be part of a group but have ulterior motives to disrupt and provoke disagreement in online discourse (Hardaker, 2015; Jachim et al., 2020), have been identified as an anti-social behavior feature in gaming groups and side discussion forums such as 4chan (Kirman et al., 2012; Samory & Peserico, 2017). Additionally, trolls may be part of government-sponsored efforts to disseminate false information about public health issues and political candidates (Broniatowski et al., 2018; Llewellyn et al., 2018; Stewart et al., 2018; Zannettou et al., 2019a), or individuals who fabricate or deceive arguments to foster mistrust and manipulate public opinion (Badawy et al., 2019). As a result, trolling is a nefarious form of online behavior that undermines public discourse and interactions, irritates conversational users, and instigates inconclusive conflict (Coles & West, 2016).

Over time, the tactics employed by trolls have evolved from utilizing aggressive speech to disseminating false news, rumors, and misinformation (Benkler et al., 2018). Additionally, trolling has transformed from sporadic actions of individuals to organized efforts by trolling farms or communities, which organize the propagation of troll content on social networking environments (SNEs) through the use of automated programs to generate misunderstandings of consensus on a specific subject (Cresci et al., 2017). Government-sponsored trolling farms utilize automated programs and fake accounts to generate troll content and manipulate community thoughts, making it challenging to detect and eliminate trolling in SNEs. These environments often rely on moderators to ban or mute trolls and flag or delete troll content as a means of circumventing this issue. However, this manual approach results in scalability challenges, delays in action, and subjectivity in judgment (Fornacciari et al., 2018; Jeun et al., 2012). To address the limitations inherent in manual methods, some detection software that operates automatically has been developed. The core functionality of these programs is based on natural language processing (NLP) algorithms and artificial intelligence (AI) approaches, such as machine learning and deep learning. As a result of the importance of automatic troll tweet detection, researchers have implemented a variety of well-established methods, such as sentimental and linguistic analysis (Capistrano et al., 2019; Fornacciari et al., 2018; Ghanem et al., 2019; Naim et al., 2022; Seah et al., 2015), social network analysis (Kumar et al., 2014; Riquelme & González-Cantergiani, 2016), and metadata analysis (Fornacciari et al., 2018; Mihaylov & Nakov, 2019).

In this study, various model architectures utilizing both NLP and deep learning techniques are developed to perform the automatic detection of troll tweets. Specifically, nine deep learning-based architectures for troll tweet detection are implemented and compared, with three models for each of the following word embedding models: BERT, ELMo, and GloVe. The performance of each architecture is evaluated in terms of precision, recall, F1 score, the area under the receiver operating characteristic curve (AUC), and classification accuracy. Results from the experiments indicate an improvement in troll tweet detection across all architectures that utilize BERT models. Furthermore, it is observed that GRU classifiers have the highest AUC score for detecting troll messages in an ELMo-based architecture.

The article is organized as follows. Section 2 provides an overview of related research on detecting troll tweets. The theoretical foundations of the various research methods employed in this study are briefly discussed in Section 3. Section 4 presents the proposed model architecture. The experiments performed and their results are outlined in Section 5. Concluding remarks are stated in the final section.

## 2. Related Work

The task of identifying and detecting troll tweets poses a significant challenge due to the pervasiveness of trolling behavior on the internet (Jachim et al., 2020). A study (Cheng et al., 2017) which investigated the antecedents of personal trolling behavior revealed that incorporating both the debate and mood context of a message can provide a more comprehensive understanding of trolling behavior than analyzing an individual's trolling history alone. A study (Llewellyn et al., 2018), linking the Russian troll farm Internet Research Agency (IRA) to state-sponsored trolling, found that a small proportion of original trolling content, such as hashtags, posts, and memes, is generated by trolls and often targeted at specific events, such as the Brexit referendum. However, a study into the activity of trolling during the 2016 US Presidential election found that Iranian trolls were opposed to Trump, while IRA trolls were pro-Trump. In an effort to evade detection, these two troll farms do not consistently employ the same tactics over time (Zannettou et al., 2019b). To facilitate constructive discourse and high levels of participation, social networking platforms aim to detect and mitigate trolling behavior, while avoiding perceptions of censorship and excessive control (Fornacciari et al., 2018). Given the potential negative impact on the reliability of public opinion and the integrity of online conversation caused by various forms and types of trolling, the need for automatic recognition of trolling in communication-based social networking platforms is evident.

To address the pervasive issue of trolling, various studies have been conducted to design algorithms for the detection of trolls. Sentiment and linguistic analysis are among the methods utilized for identifying trolls as the contents of their messages are primarily textual and often contain provocative and inflammatory language. In (Seah et al., 2015), a domain-adapting sentimental analysis was developed to differentiate trolling activities by utilizing linguistic, thread-level, user-level, and post-level features. Results yielded an approximate detection rate of 70% in an online forum. In addition, the measurement of emotions and sentiments in messages can aid in detecting trolling on Twitter. The authors of (Jachim et al., 2020) achieved a 76% detection rate of Twitter trolling by evaluating abusive textual content along with other meta-data from troll messages. In (Hutto & Gilbert, 2014), the authors attained an accuracy of 88.95%, recall of 93.12%, and precision of 86.88% by implementing an out-of-box sentimental analysis tool named VADER in conjunction with syntactic, lexical, and aggression analyzers, using the Kaggle Twitter cyber-trolls dataset (Capistrano et al., 2019; DataTurks, 2022). In (Ghanem et al., 2019), a troll tweet detection algorithm analyzing the IRA Twitter writing style trolling was employed, resulting in an F1 score of 94%. A significant amount of metadata about users is also provided by online and social media platforms, which can be utilized to detect and understand the behavior of trolls. In (Mihaylov & Nakov, 2019), the authors generated two classifiers, utilizing sentimental analysis and data regarding the timing of troll messages (weekend, workday, nighttime, work time), with the first classifier attempting to detect sponsored trolls and the second aiming to detect individualistic trolling. Results yielded parallel detection rates of around 82% accuracy. In another study (Kumar et al., 2014), the authors obtained a precision of 51% using metadata in a broader investigation of the Slashdot Zoo website, discriminating against troll users. To distinguish real individuals from state-sponsored trolls and individual trolls, the authors of (Riquelme & González-Cantergiani, 2016) implemented a different method to analyze the activity and effects of users on Twitter. The authors of (Miao et al., 2020) propose an approach for detecting troll tweets in English and Russian by reducing the detection process to authorship verification. The proposed method is evaluated using monolingual, cross-lingual, and bilingual training scenarios with multiple machine learning algorithms, such as deep learning. Bilingual learning achieved the best results with AUCs of 0.877 and 0.82 for tweet classification in English and Russian test sets, respectively.

3. Background
3.1. Word Embedding Methods

### 3.1.1. GloVe

To generate word embeddings, GloVe (Global Vectors for Word Representation) is an alternative technique (Mujtaba, n.d.). The GloVe word representations are a mixture of prediction approaches and count-based approaches such as Word2Vec. The GloVe stands for Global Vectors (Voita, n.d.-a), which represents the idea: that the technique employs global data from a corpus to acquire word vector representations (Voita, n.d.-b). To calculate the relation between context C and word W: N (W, C), the easiest count-based technique employs co-occurrence counts. To build the loss function, Global Vectors utilizes such counts as well. GloVe is based on matrix factorization methods on the word-context matrix. A big matrix of co-occurrence information is built, and each 'word' (the rows) is counted, and how often this word in some 'context' (the columns) is seen in a huge corpus. Generally, the corpus is scanned in the following way: context terms are searched for in a specified range set by a window size for every term before the term and a window size after the term. In addition to this, a smaller amount of weight is given to more distant words. The number of contexts is huge because this is fundamentally combinatorial in size. Consequently, to produce a lower-dimensional matrix at that time, this matrix is factorized, where for each word each row produces an embedding vector. It is generally completed by making a reconstruction loss minimum. Discovering the lower-dimensional word embeddings that can describe many variations in the high-dimensional information is the aim of this loss.

### 3.1.2. ELMo

A novel sort of deep contextualized word representation that models both (1) complex features of word usage (e.g., semantics and syntax), and (2) how these employs vary across linguistic contexts such as modeling polysemy (Peters et al., 2018) are introduced. ELMo is different from conventional types of word embedding in that every token is assigned a representation that is the complete input sentence function. In this model, a bi-directional LSTM that is trained by a coupled language model objective on a huge amount of text corpora generates vectors. Because of this, it is called ELMo (Embeddings from Language Models) representations. ELMo representations, in contrast to earlier methods for contextualizing word vectors (McCann et al., 2017; Peters et al., 2017), are deep, in the sense that they are a function of the entire internal layers of the bidirectional language model. For each end task, more particularly, a linear vector combination is stacked above every input word, which increases performance over just utilizing the top LSTM layer learned. In this way, the internal state's combination permits very rich word representations. It has been proven that context-dependent aspects of word meaning (for example they can be utilized without alteration to make well performance on supervised word sense disambiguation tasks) are captured by the higher-level LSTM states whereas lower-level states model aspects of syntax (for example the part-of-speech tagging is done by them). Simultaneously revealing these signals is extremely beneficial, as it enables the learned models to select the semi-supervision techniques that are most advantageous for each final task.

### 3.1.3. BERT

Google introduced Bidirectional Encoder Representations from Transformers (BERT), which is a language transformation method (Devlin et al., 2018). The BERT learns the deep representation of texts by taking into account both, right and left contexts. For that reason, it is called deeply bidirectional. BERT is a technique that can be utilized for training a general-purpose language model on very big corpora and then these models can be used for Natural Language Processing (NLP) tasks (Bedi & Toshniwal, 2022; Maslej-Krešňáková et al., 2020; B. Zhou et al., 2022). Consequently, pre-training and fine-tuning are two different phases used when implementing BERT. is trained on unlabeled data is used to train during the pre-training step. After that, the pre-trained parameters are used to initialize the framework and then the framework will fine-tune for certain NLP tasks.
Enormous computation resources are required to fine-tune the BERT framework. The identical framework is used by BERT in various tasks. The Transformers is used to build the BERT framework (Vaswani et al., n.d.). The framework has two variations, which are the BERT-large and BERT-base models. 16 self-attention heads, a hidden size of 1024, and 24 Transformer blocks are included in BERT-large and hidden size of 768, 12 self-attention heads and 12 Transformer blocks are included in the BERT-base model.

### 3.2. Encoder Methods
### 3.2.1. CNN

A particular sort of feed-forward neural network is represented by CNN (Convolutional neural networks), which includes neurons of a layer performing a convolution operation (Maslej-Krešňáková et al., 2020). CNN networks are inspired by the ocular nerve function. Following a specific size of the convolution filter, neurons react to the activations of the input of the nearby neurons named filters as well. Shifting the convolutional filter over the entire set of values is the process of convolution. In such a situation, input values and the product of the convolution filter are shown by the convolution process (Goodfellow et al., 2016). To decrease the number of outputs, and computational complexity and to prohibit overfitting in networks, pooling layers are used in CNN. Immediately after the convolution layers, the sampling layers are generally processed since when the convolution filters are shifting throughout the single inputs, the duplicated values are produced. The pooling layers are used to remove surplus data. The global-max pooling layers where we select between the global max-pooling layer and the global-average pooling layer are used in this study. Such layers perform on the same principle as the conventional max-pooling layer. The dissimilarity is that average pooling is calculated for the complete input, not just a specific field.

### 3.2.2. RNN

Recurrent neural network (RNN) is a traditional sort of neural network with recurring connections, which permits a type of memory (J. Zhou et al., 2019). In addition, these make it appropriate for sequential prediction tasks that contain arbitrary spatiotemporal dimensions. Therefore, various NLP problems employ the RNN architecture to analyze sequence tokens by concerning a sentence interpretation.

Long Short-Term Memory (LSTM) (Hochreiter & Schmidhuber, 1997) is a sort of RNN. To solve the vanishing gradient problem in RNN, LSTM contains a more complicated framework. Therefore, LSTM will make certain that long-term context and data are preserved in any sequence-based problems. In comparison with other neural network architectures, instead of interconnected neurons, the LSTM structure includes memory blocks that are linked in layers. To deal with the data flow, the block output, and the block state, the LSTM uses gateways in the block. Gateways can decide which information should be kept and which is significant in a sequence. LSTM has four gates, which perform the functions below: i) Input gate—this controls the entry of data into the memory block. ii) Cell state—long-term data is stored in this gate. iii) Forget gate— this gate can learn what data will be preserved and what data will be removed and then perform actions according to that. iv) Output gate—this gate will make a decision on what action to carry out on the output according to the memory unit and the input. In addition to that, the authors of (Cho et al., 2014) aim to resolve the vanishing gradient problem, where they present a type of RNN model called Gated Recurrent Unit (GRU). GRU can be thought of as a type of LSTM model since the design of both networks is done in a parallel way. To deal with the vanishing gradient problem, GRU uses reset and update gates. The reset gate aids the network to decide, how much data will be deleted from the prior time steps and the update gate assists the network to decide, how much of the prior data from the earlier time steps will be required in the next steps.

### 3.2.3. Transformer

Nowadays, transformer models are very common techniques used to resolve a variety of natural language problems for example language understanding, summarization, and question answering but especially it is effectively employed in text categorization problems (Sun et al., 2019). Similar to RNN structure, sequential information for example natural language tasks such as text summarization and translation (*Transformer (Machine Learning Model) - Wikipedia*, n.d.) is dealt with by Transformers. On the other hand, in comparison with recurrent neural networks, processing the sequential data in order is not needed in Transformers. For instance, if we have a natural language sentence, the start of this sentence is not required to be processed before the end in the Transformer model. Because of this characteristic, the Transformers permit considerably more parallelization in comparison with recurrent neural networks and as a result decrease training times (Vaswani et al., n.d.). In NLP, Transformers have quickly turned out to be the architecture of choice (Wolf et al., 2020) by substituting previous RNN structures for instance LSTM.

A transformer is an encoder-decoder model. In the encoder part, some encoding layers repetitively process the input data one after the other. In the decoder part, some decoding layers perform the identical thing to the encoder output. Processing the input to produce encodings, which contain data regarding which input parts are related to each other, is the purpose of every encoder layer. The encoders pass their encodings set to the following encoder layers as input.

To produce output sequences, each decoder will take the entire encodings and process them via their incorporated contextual data. In other words, each decoder layer performs the opposite of the encoder (*Sequence Modeling with Neural Networks (Part 2): Attention Models – Indico Data*, n.d.). To accomplish this, an attentional mechanism is used by each decoder and encoder layer, which for every input, measures every other input relevance and takes data from them accordingly to generate the output (Alammar, n.d.-b). In addition, all decoder layers contain an extra attentional mechanism taking data from the prior decoder's outputs before the decoder layer takes data from the encodings. For extra operating of the outputs, a feed-forward neural network is included in both the decoder and encoder layers. Additionally, they have layer normalization phases and residual connections (Alammar, n.d.-b).

### 3.2.4. Evaluation Metrics

There are four categories for the results of binary classification (Fawcett, 2004): i) True Positive ($TP$): Classifying positive samples correctly; ii) False Negative ($FN$): Classifying a positive sample inaccurately; iii) False Positive ($FP$): Classifying a negative sample inaccurately; VI) True Negative ($TN$): Classifying negative samples correctly; as a starting point, following metrics can be obtained via the previous ones as well.

For classification, the most fundamental evaluation measure is accuracy. The fraction of the accurate predictions is represented by the evaluation measure accuracy in the network. The calculation of accuracy is as follows:

$$Accuracy = (TP + TN)/(TP + FP + TN + FN)$$

The fraction of accurate predictions between the entire predictions with the positive label is the precision, which represents how many of the positively predicted samples are true positive samples. The calculation of the standard precision is shown below:

$$Precision = TP/(TP + FP)$$

Between all positive samples, the fraction of accurate predictions is the recall, which indicates how many positive samples are positively classified. The calculation of the standard recall is shown below:

$$Recall = TP/(TP + FN)$$

The harmonic average of recall and precision represents the F-measure, which can be calculated as follow:

$$F1 = 2\frac{Recall \times Precision}{Recall + Precision}$$

Here, the weight of recall and precision is equal. For that reason, this is called an F1 measure as well.

Receiver Operating Characteristics (ROC): ROC curve is selected as a standard criterion to test classifiers if there is a dataset with a class imbalance issue (He & Ma, 2013; Tharwat, 2018). To evaluate a classifier's performance, the area under the ROC curve (AUC) metric is commonly utilized as a de facto measure if a dataset has a class imbalance issue. The AUC is a metric, which calculates how much a positive class sample is higher ranked than a negative class sample randomly when the samples are sorted with classification probabilities.

## 4. The proposed model architecture

To enable the processing and analysis of text such as Twitter messages, NLP methods such as machine translation models and language models are frequently employed. The initial step in determining the meaning of a text is to acquire an appropriate representation of the text. When treating each word or sub-word as a distinct token, pre-trained representations of individual words may be obtained through the use of various techniques, including GloVe, ELMo, and BERT. After completing the pre-training process, each word is then represented as a vector. However, some techniques, such as GloVe and FastText, consistently utilize the same word vector representation for a given word regardless of the context in which it appears. Conversely, various pre-trained methods modify the word vector representation of the same token to accommodate different contexts. These methods provide word embeddings based on the context in which the word is used, thereby enabling the determination of the word's meaning in that particular context (Alammar, n.d.-a). BERT is a self-supervised method that is much deeper than other methods and is based on the transformer encoder. In this research, different pre-trained word representations will be employed to evaluate the troll tweet dataset.

Pre-trained word embeddings can be fed to many deep neural network models for a variety of NLP tasks. As can be seen from Figure 1, the many pre-trained word embedding methods can be used to feed a range of deep neural network models for diverse NLP problems.

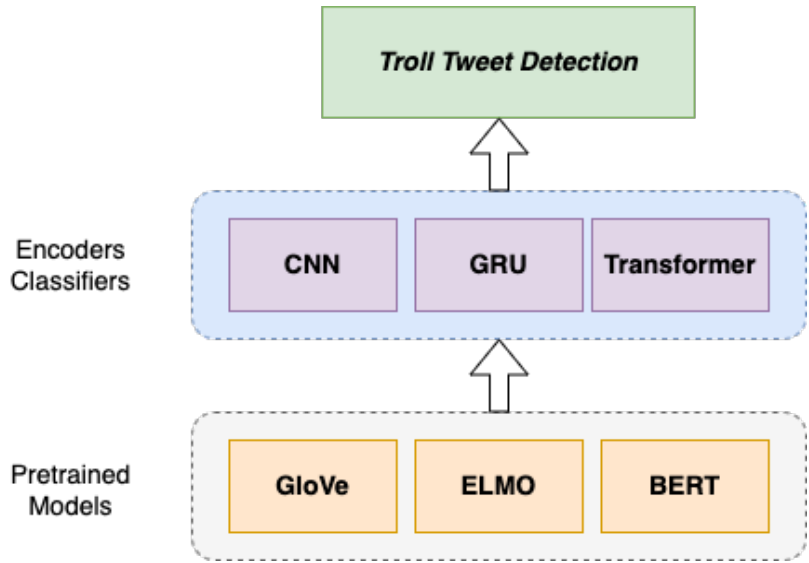

**Figure 1.** The proposed model architecture

Detecting troll tweets is one of the fundamental tasks of NLP, which involves evaluating the category of a sentence containing a bunch of words. Troll tweet detection mainly consists of two parts:

1) Word embedding, for example, uses the classic BERT, ELMo, and GloVe, which have been popular in recent years, to convert words into word vectors.
2) The encoder performs the next feature extraction of word embedding inputs, such as the commonly used CNN, GRU and Transformer models that have been very popular in recent years, which are more high-level semantic features that can extract words.

This study focuses on these two parts, testing the effects of text classification in different word vectors with different encoder.

## 5. Experiments

### 5.1. Dataset Description

Experiments are conducted using a troll tweet dataset compiled by Miao et al. (Miao et al., 2020). This dataset consists of 18,514 trolls collected from 2018 to 19. Each message is labeled with one of the following two classes: Troll (9257 messages), and non-troll (9257 messages). In addition, duplicates and retweets are filtered out to prevent over-fitting in this dataset (Miao et al., 2020). The same number of normal tweet messages is gathered at random to generate a balanced dataset for binary classification. The same hashtags as those of the troll messages are utilized by the list of random messages. The ultimate troll tweet dataset includes 18,514 English tweet messages, after adding tweet messages from real accounts. The dataset is divided into three parts so that it contains 12.959 training samples, 1.851 validation samples, and 3.702 test samples. The sample number distribution reveals that this data set is well-balanced.

### 5.2. Experimental setup

The following word embedding vector types were chosen for this study: GloVe utilizes the word vector Glove.6B.300d. ELMo employs the context word vector of the pre-trained model supplied by AllenNLP; the dimensions of the LSTM's implicit and output layers are 2048/256. BERT employs the Bert-large-uncased word vector supplied by Transformers, which has a 12-layer transformer structure and a 512-byte output layer size.
Following are descriptions of three distinct encoder methods. CNN utilizes convolutional neural networks for feature extraction. RNN extracts features utilizing a GRU network. The transformer was combined with Positional Embedding and Transformer Encoder to extract features. The implementation of PyTorch was utilized for this procedure.

### 5.3. Performance Evaluation

The proposed models have been evaluated in terms of F1 measure and classification accuracy on the dataset explained above in the experiment. If the categories of the dataset are balanced, the accuracy metric is one of the best evaluation metrics in terms of performance (Singh et al., 2022). But the troll tweet dataset used in this study has a balance over its classes. Although the classifier's performance can be validated by the accuracy metric only, high values for the F1 score along with a high AUC score have been tried to accomplish in addition to the good accuracy score in the experiment.

Table 1. Performance results in terms of accuracy, precision, recall, F1, and AUC

| Pretrained Embedding Method | Encoder Methods | Accuracy | Precision | Recall | F1 | AUC |
|---|---|---|---|---|---|---|
| BERT | CNN | 0.838 | 0.849 | 0.822 | 0.835 | 0.915 |
|  | GRU | 0.845 | **0.865** | 0.817 | 0.840 | 0.924 |
|  | Transformer | **0.856** | 0.825 | **0.904** | **0.863** | 0.909 |
| ELMo | CNN | 0.842 | 0.833 | 0.854 | 0.844 | 0.916 |

|  | GRU | 0.855 | 0.839 | 0.878 | 0.859 | **0.929** |
|  | Transformer | 0.843 | 0.827 | 0.867 | 0.847 | 0.917 |
| **GloVe** | CNN | 0.743 | *0.743* | 0.741 | 0.742 | 0.818 |
|  | GRU | 0.767 | 0.760 | 0.779 | 0.769 | 0.831 |
|  | Transformer | *0.732* | 0.749 | *0.698* | *0.723* | *0.806* |

*Bold values indicate the best-performed method and bold italics indicate worst performed method.*

It is vital to use decent embedding methods to preprocess the specified dataset to achieve excellent troll tweet detection performance (Guo et al., 2019). Nowadays, the word embedding methods such as BERT, GloVe, and ELMo are good selections.

The results of the experiment, as presented in Table 1 and Table 2, indicate that the BERT and ELMo embedding methods perform better than the GloVe method in detecting troll tweets in the dataset. This finding is consistent with the literature, which suggests that contextualized word embeddings, such as those generated by BERT and ELMo, provide a more accurate representation of a word's meaning in a given context, compared to traditional word embedding methods such as GloVe, which provide the same word vector representation for the same word regardless of the context.

The use of contextualized word embeddings, such as those provided by BERT and ELMo, is particularly important in the task of troll tweet detection, as it allows the models to take into account the nuances and subtleties of language use in the context of online social media platforms, where the meaning of words can be highly dependent on the context in which they are used. This is important because troll tweets often contain manipulative language in order to incite reactions from other users, and a model that is not able to take into account the context in which words are used is likely to miss these nuances and make errors in classification.

Table 2. Comparison of the AUC results with the previous study

| Ref. | Method | AUC |
|---|---|---|
| Miao et al. (Miao et al., 2020) | SVM (Stylometric) | 0.836 |
|  | Logistic Regression (N-grams) | 0.861 |
|  | RNN+CNN (Raw Text) | 0.824 |
| *This study* | BERT+CNN | 0.915 |
|  | BERT+GRU | 0.924 |
|  | BERT+Transformer | 0.909 |
|  | ELMo+CNN | 0.916 |

|  | **ELMo+GRU** | **0.929** |
|  | ELMo+Transformer | 0.917 |
|  | GloVe+CNN | 0.818 |
|  | GloVe+GRU | 0.831 |
|  | GloVe+Transformer | *0.806* |

In addition, the results show that the CNN, GRU, and transformer encoder methods performed similarly in terms of F1 score and AUC. This suggests that these architectures have similar capabilities in extracting relevant information from the input text for the task of troll tweet detection. These architectures have been widely used in various NLP tasks and have been proven to perform well in tasks such as text classification, sentiment analysis, and text generation. The results of this experiment are consistent with the literature in this regard, which suggests that these architectures are effective at extracting important features from the text and performing well on a wide range of NLP tasks.

Furthermore, the results demonstrate that the BERT and ELMo embedding methods performed better when combined with the CNN and GRU encoders than with the transformer encoder. This is likely due to the fact that the CNN and GRU encoders are particularly well suited for extracting relevant information from the contextualized word embeddings provided by BERT and ELMo. In contrast, the transformer encoder, which is primarily designed for sequence modeling tasks, may not be as effective at extracting the relevant information from the contextualized word embeddings for the task of troll tweet detection. This finding is in line with the literature, which suggests that different architectures are suitable for different NLP tasks and that the choice of architecture is highly dependent on the specific characteristics of the task at hand.

The results in Table 2 also show that the ELMo+GRU method has a better AUC score compared to previous studies. The AUC scores of these methods outperform the state-of-the-art results of other methods. The AUC is considered a robust evaluation metric as it takes into account the false positive rate and true positive rate simultaneously. This means that a high AUC value indicates that the false positive rate is low and the true positive rate is high, which makes the model a reliable one in detecting troll tweets. The high AUC value of 0.929 for these methods suggest that these models have a low false positive rate and a high true positive rate, which makes them reliable for troll tweet detection.

In summary, the results of the experiment indicate that the BERT and ELMo embedding methods, when combined with the CNN and GRU encoder methods, are effective at detecting troll tweets in the dataset. The use of contextualized word embeddings provided by BERT and ELMo, in combination with the CNN and GRU encoder methods, allowed the models to extract relevant information from the input text and achieve high performance in terms of F1 score and AUC. The results are consistent with the literature and demonstrate the effectiveness of these methods in detecting troll tweets on online social media platforms.

## 6. Conclusion

In recent years, the proliferation of trolling accounts on social media has become a major concern, as these accounts can manipulate public opinion and create a hostile online environment. As such, the task of detecting troll messages and taking appropriate action is of crucial importance for maintaining a safe and healthy social media environment.

In this study, we proposed a set of nine different model architectures for troll tweet detection, which were based on deep learning. These models made use of three different pre-trained word embedding methods: BERT, ELMo, and GloVe. The performance of each architecture was evaluated using a number of metrics, including classification accuracy, F1 score, AUC, and precision.

The results of our experiments indicate that the BERT and ELMo embedding methods performed better than the GloVe method in detecting troll tweets in the dataset. This can likely be attributed to the fact that these methods provide contextualized word embeddings, which allows the model to take into account the nuances and subtleties of language use in the context of online social media. Additionally, in terms of the encoder methods, we found that the CNN and GRU encoders performed similarly in terms of F1 score and AUC, suggesting that these architectures have similar capabilities in extracting relevant information from the input text for the task of troll tweet detection. The best-performed method in terms of AUC is the customized ELMo-based architecture that employs a GRU classifier with a score of 0.929. This result highlights the importance of utilizing contextualized word embeddings and appropriate encoder methods in the task of troll tweet detection, which can be used by various social-based systems to improve their performance in identifying and addressing trolling behavior on their platforms.

However, there are still many open research questions to be addressed in the field of troll tweet detection. For example, handling long texts, identifying trolls' behavioral characteristics, and utilizing advanced data mining techniques to improve troll tweet detection performance are all important areas for future research. Additionally, it would be important to study the generalization performance of the proposed models to different datasets and languages. Furthermore, we need more labeled datasets to evaluate the performance of the proposed models on more realistic datasets. Therefore, this study is only a step forward in the field of troll tweet detection and more research is needed to improve the performance of the proposed models.

**Conflict of interest**

The authors declare that they have no conflict of interest.